\documentclass[journal,twoside,web]{ieeecolor}
\usepackage{generic}
\usepackage{caption}
\usepackage{url}
\usepackage{subcaption}
\usepackage{pifont}
\usepackage{threeparttable}
\usepackage{amsmath,amssymb,amsfonts}
\usepackage{algorithmic}
\usepackage{graphicx}
\usepackage{algorithm,algorithmic}
\usepackage{multirow}
\usepackage{textcomp}
\usepackage{titlesec}
\usepackage{orcidlink}
\let\labelindent\relax
\usepackage{enumitem}
\usepackage{wrapfig}
\titlespacing*{\subsection}{0pt}{2pt}{2pt}
\usepackage{float}

\usepackage{hyperref}
\hypersetup{
    colorlinks=true,
    citecolor=blue,
    linkcolor=blue,
    urlcolor=blue,
    bookmarksnumbered=true,
    bookmarksopen=true,
    hidelinks
}

\def\BibTeX{{\rm B\kern-.05em{\sc i\kern-.025em b}\kern-.08em
    T\kern-.1667em\lower.7ex\hbox{E}\kern-.125emX}}

\begin{document}
\title{Accurate Cobb Angle Estimation via SVD-Based Curve Detection and Vertebral Wedging Quantification}

\author{Chang Shi\textsuperscript{\dag},
Nan Meng\textsuperscript{\dag}*, \textit{IEEE Member},
Yipeng Zhuang\textsuperscript{\dag}, \textit{IEEE Member},
Moxin Zhao,
Hua Huang,
Xiuyuan Chen,
Cong Nie,
Wenting Zhong,
Guiqiang Jiang,
Yuxin Wei,
Jacob Hong Man Yu,
Si Chen,
Xiaowen Ou,
Jason Pui Yin Cheung,
Teng Zhang*, \textit{IEEE Senior Member}
\thanks{\textsuperscript{\dag}C. Shi, N. Meng, and Y. Zhuang contributed equally to this work as co-first authors.}
\thanks{This work was supported in part by National Natural Science Foundation of China Young Scientists Fund(Grant ID: 82402398, 82303957).}\thanks{Corresponding authors: *T. Zhang, N. Meng.}
\thanks{C. Shi, N. Meng, Y. Zhuang, M. Zhao, H. Huang, C. Nie, W. Zhong, G. Jiang, Y. Wei, J. H. M. Yu, S. Chen, X. Ou, J. P. Y. Cheung, and T. Zhang are with the Department of Orthopaedics and Traumatology, LKS Faculty of Medicine, The University of Hong Kong, Hong Kong (e-mail: chaseshi@hku.hk; nanmeng@hku.hk; yipengzh@hku.hk; moxin@connect.hku.hk;\linebreak huangh246@mail2.sysu.edu.cn; niecong@hku.hk; wtzhong@hku.hk;\linebreak fcostars@hku.hk; u3012166@connect.hku.hk; jacobyu@connect.\linebreak hku.hk;  cyn08ccs@hku.hk; xiaoweno@hku.hk; cheungjp@hku.hk; tgzhang@hku.hk).}
\thanks{X. Chen is with the Department of Spine Surgery, Renji Hospital, School of Medicine, Shanghai Jiao Tong University, Shanghai, China (e-mail: chenxiuyuan@renji.com).}
}

\maketitle

\vspace{-4pt}
\noindent\textit{\footnotesize Accepted for publication in IEEE Journal of Biomedical and Health Informatics}
\vspace{-2pt}

\begin{abstract}
Adolescent idiopathic scoliosis (AIS) is a common spinal deformity affecting approximately 2.2\% of boys and 4.8\% of girls worldwide. The Cobb angle serves as the gold standard for AIS severity assessment, yet traditional manual measurements suffer from significant observer variability, compromising diagnostic accuracy. Despite prior automation attempts, existing methods use simplified spinal models and predetermined curve patterns that fail to address clinical complexity. We present a novel deep learning framework for AIS assessment that simultaneously predicts both superior and inferior endplate angles with corresponding midpoint coordinates for each vertebra, preserving the anatomical reality of vertebral wedging in progressive AIS. Our approach combines an HRNet backbone with Swin-Transformer modules and biomechanically informed constraints for enhanced feature extraction. We employ Singular Value Decomposition (SVD) to analyze angle predictions directly from vertebral morphology, enabling flexible detection of diverse scoliosis patterns without predefined curve assumptions. Using 630 full-spine anteroposterior radiographs from patients aged 10-18 years with rigorous dual-rater annotation, our method achieved 83.45\% diagnostic accuracy and 2.55° mean absolute error. The framework demonstrates exceptional generalization capability on out-of-distribution cases. Additionally, we introduce the Vertebral Wedging Index (VWI), a novel metric quantifying vertebral deformation. Longitudinal analysis revealed VWI's significant prognostic correlation with curve progression while traditional Cobb angles showed no correlation, providing robust support for early AIS detection, personalized treatment planning, and progression monitoring.
\end{abstract}
\vspace{-8pt}
\vspace{-6pt}
\section{Introduction}

\begin{figure}[!t]
    \centering
    \includegraphics[width=0.8\columnwidth]{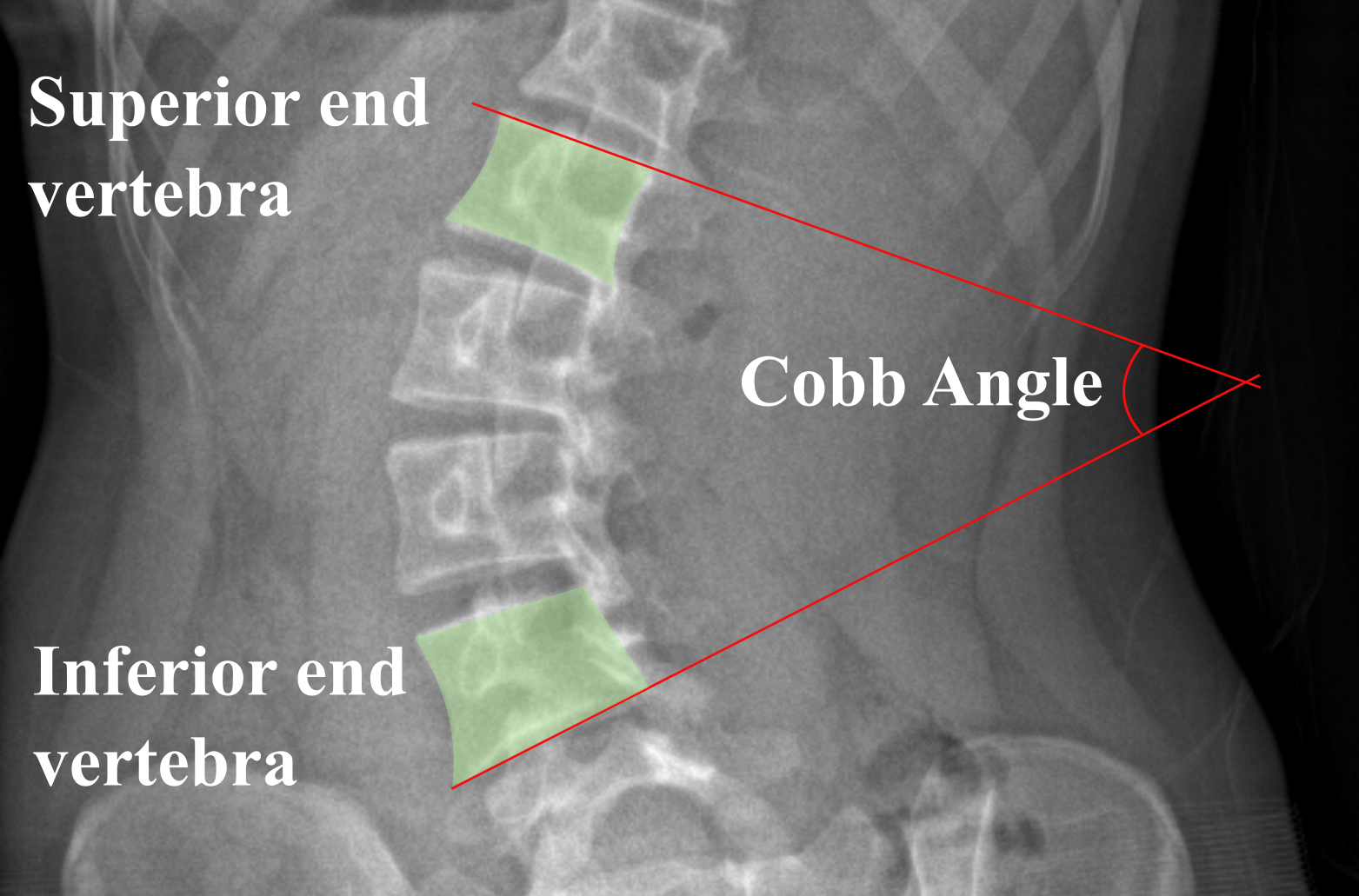}
    \caption{Illustration of Cobb angle measurement on a posteroanterior radiograph.}
    \label{fig:cobb_angle}
\end{figure}

Adolescent idiopathic scoliosis (AIS) is the most prevalent spinal deformity affecting adolescents, with approximately 2.2\% of boys and 4.8\% of girls worldwide suffering from this condition \cite{horne2014adolescent, weinstein2008adolescent}. If left untreated during early developmental stages, AIS can significantly disrupt physical development, potentially leading to pronounced postural deformities and, in severe cases, compromised cardiopulmonary function \cite{weinstein2019natural,meng2022artificial}. In current clinical practice, standing posteroanterior (PA) spinal radiographs are routinely used by physicians to diagnose and evaluate spinal curvature. The Cobb angle remains the gold standard for quantitatively assessing the severity of spinal deformity \cite{fong2015population, cassar2002imaging,zhang2023deep}. As illustrated in Fig.~\ref{fig:cobb_angle}, this measurement is obtained by identifying the upper and lower end vertebrae with the greatest inclination and measuring the angle between lines drawn along the superior endplate of the upper end vertebra and the inferior endplate of the lower end vertebra. In cases presenting multiple scoliotic curves, disease severity is defined by the largest Cobb angle observed.

Despite its widespread clinical use, manual Cobb angle measurement remains subject to considerable variability, with inter-observer differences ranging from 5° to 10° \cite{jin2022review}. Such variability can lead to missed diagnoses of early progressive cases or overtreatment, subsequently affecting patient prognosis and treatment outcomes. Furthermore, manual measurement methods are time-consuming and dependent on expert experience, making them inadequate for large-scale screening and continuous monitoring \cite{kumar2024critical}. Consequently, there is a pressing need for a high-precision, automated Cobb angle measurement system to enable early detection and personalized treatment of AIS.

Existing automated Cobb angle measurement methods encounter three primary limitations. First, insufficient clinical interpretability. Some methods directly obtain Cobb angles or classification results from radiographs \cite{liu2023lenke, lin2020seg4reg, horng2019cobb}, omitting intermediate anatomical landmarks and diverging from clinical workflow. Second, simplified vertebral morphology. Some methods \cite{yao2022w, zou2023vltenet, zhao2023spinehrformer} simplify vertebrae to a single tilt angle, failing to reflect vertebral complexity. As AIS progresses, vertebrae develop wedge-shaped deformities \cite{wever1999biomechanical, noshchenko2015predictors} with non-parallel endplates, which simplified modeling approaches cannot accurately capture. Third, inability to adapt to diverse curve patterns. Detection methods based on predetermined curve structures \cite{zou2023vltenet, lin2020seg4reg, chen2019accurate} cannot identify the complex patterns encountered in clinical practice \cite{lenke2003lenke, slattery2018classifications}.

Recent deep learning methods have advanced AIS assessment. Segmentation-based approaches \cite{liu2023lenke, lin2020seg4reg} extract morphological features through vertebral segmentation for Lenke classification and Cobb angle regression, while visually intuitive, they lack clinical interpretability. Recent advances have improved robustness: Suri et al. \cite{suri2023conquering} developed hardware-invariant algorithms for spinal instrumentation cases, Kato et al. \cite{kato2024comparison} enhanced accuracy through cross-disease training, and Wang et al. \cite{wang2024deep} achieved efficient processing with competitive precision. Horng et al. \cite{horng2019cobb} applied minimum bounding rectangles post-segmentation but neglected internal vertebral structures. Landmark-based methods have shown improvements: W-Transformer \cite{yao2022w} improved global feature extraction using transformers, while Zou et al. \cite{zou2023vltenet} combined positional heatmaps with vector maps to predict vertebral tilt angles. However, these methods still rely on fixed curve pattern assumptions and simplify vertebrae to single tilt angles, ignoring non-parallel endplates in progressive AIS. SpineHRformer\cite{zhao2023spinehrformer} enhanced curve recognition using expert-labeled end vertebrae but introduced scalability challenges. These limitations can lead to inaccurate measurements in complex cases and missed clinically significant curves.

To overcome the limitations of existing methods, we propose a novel deep learning framework for automated assessment of AIS with the following key contributions:
\begin{itemize}[leftmargin=1em]

    \item \textbf{Dual-task for Vertebral Morphology Preservation:}

    We design a dual-task framework to simultaneously regress vertebral endplate center locations and their inclination angles. This approach preserves anatomical details and models wedge-shaped vertebral deformities in progressive AIS. Furthermore, we introduce Vertebral Wedging Index (VWI), a novel metric quantifying vertebral deformation within scoliotic curves. Through longitudinal analysis of 138 patients, VWI demonstrates superior prognostic value with significant correlation to curve progression.

    \item \textbf{Biomechanically Informed Loss:}

    We incorporate prior anatomical knowledge directly into model training through a biomechanically informed loss function. By constraining spatial relationships between adjacent vertebrae, this loss enhances the anatomical plausibility of model outputs. Additionally, we integrate Swin-Transformer modules to strengthen long-range dependency modeling and improve prediction accuracy.

    \item \textbf{SVD-based Principal Curve Detection with Validation:}

    We propose an SVD-based method to extract principal curvature patterns directly from predicted endplate angles, without assuming fixed curve numbers or predefined anatomical regions. This approach enables flexible detection of diverse scoliosis patterns with exceptional generalization capability, achieving 78.8-82.7\% agreement with expert evaluators on 104 out-of-distribution cases.
\end{itemize}

\section{Methodology}

\begin{figure*}[!t]
    \centering
    \includegraphics[width=0.9\textwidth]{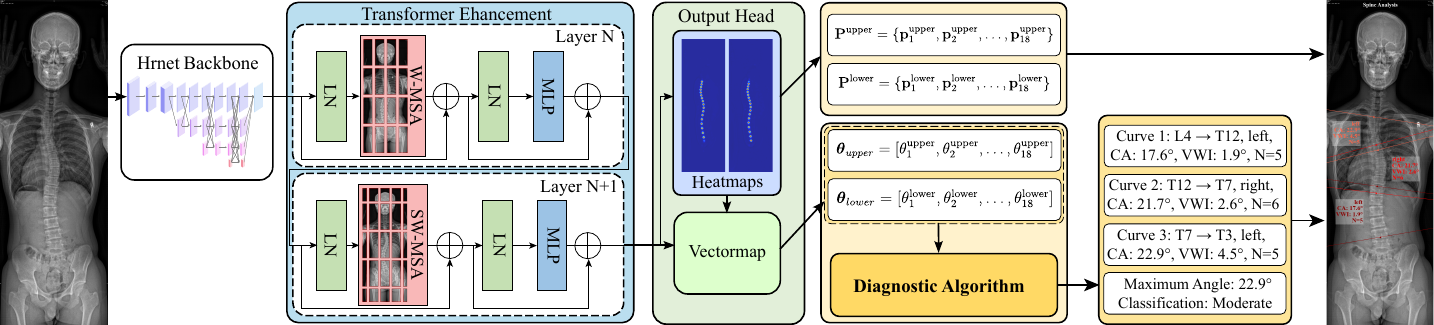}
    \caption{Overview of the AIS assessment framework. Input spinal radiographs are processed by an HRNet backbone for feature extraction and Swin Transformer layers with W-MSA and SW-MSA blocks to enhance global context. The dual-task output head simultaneously predicts the positions and inclination angles of each vertebra's upper and lower endplates. A dual-task head predicts vertebral landmarks and inclinations, followed by WVI and Cobb angle computation.}
    \label{fig:framework_pipeline}
\end{figure*}
\subsection{Problem Definition}
Given a radiographic spinal image $\mathbf{I} \in \mathbb{R}^{H \times W}$, the objective is to predict, for each vertebra, both the superior and inferior endplate angles relative to the horizontal axis, which more accurately reflects the clinical gold standard for Cobb angle measurement.

Let $\mathcal{V} = \{v_1, v_2, ..., v_N\}$ represent the set of $N$ vertebrae in the spine, where each vertebra $v_i$ has a superior endplate with angle $\theta_i^{\text{upper}}$ and an inferior endplate with angle $\theta_i^{\text{lower}}$ relative to the horizontal axis. These angles are defined in the range $[-90^{\circ}, 90^{\circ}]$, where $0^{\circ}$ represents parallel to horizontal orientation. In addition, we present the midpoint coordinates of each endplate, denoted as $\mathbf{p}_i^{\text{upper}} = \left(x_i^{\text{upper}}, y_i^{\text{upper}}\right)$ and $\mathbf{p}_i^{\text{lower}} = \left(x_i^{\text{lower}}, y_i^{\text{lower}}\right)$.

Our deep learning model $f$ with parameters $\Theta$ therefore aims to learn the mapping:
\begin{equation}
f_{\Theta}: \mathbf{I} \mapsto \left\{ \left(\theta_i^{\text{upper}}, \theta_i^{\text{lower}}, \mathbf{p}_i^{\text{upper}}, \mathbf{p}_i^{\text{lower}}\right)\right\}_{i=1}^N.
\end{equation}

The overall AIS assessment process is shown in Fig.~\ref{fig:framework_pipeline}.

\subsection{Dataset}
\begin{figure}[!t]
    \centering
    \includegraphics[width=0.8\columnwidth]{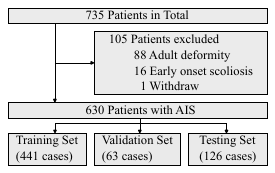}
    \caption{Flowchart of patient selection and dataset division.}
    \label{fig:patient_flowchart}
\end{figure}

We conducted a dual-center retrospective study collecting data from October 9, 2019, to March 19, 2025, comprising patients with AIS aged between 10 and 18 years. Patients were recruited from two tertiary medical centers in Hong Kong: Queen Mary Hospital and The Duchess of Kent Children's Hospital.

At both centers, whole-spine standing posteroanterior radiographs were acquired using standardized protocols with the EOS™ (EOS® Imaging, Paris, France) biplanar stereoradiography system. All radiographs were independently annotated by two experienced orthopedic surgeons with over 15 years of spine surgery experience, following the same protocol established in our previous studies~\cite{meng2022artificial,zhang2023deep}. For annotation consistency across both centers, each of the 18 vertebrae (from the 7th cervical vertebra to the 5th lumbar vertebra) was marked with four landmarks corresponding to the left and right endpoints of both the superior and inferior endplates.

Inter-rater reliability analysis showed a mean absolute difference of 3.1° ± 1.8° for Cobb angle measurements, with 91.5\% of cases demonstrating agreement within ±5°. Cases with disagreement \textgreater 5° (n=32, 4.4\%) underwent joint consensus review by both raters.

From our combined 735 annotated cases, we excluded 105 cases (88 with adult spinal deformity, 16 with early-onset scoliosis, and 1 withdrawal), resulting in 630 eligible AIS cases for analysis. We divided this dataset using a 7:1:2 ratio: 441 cases for training, 63 for validation, and 126 for testing, as illustrated in Fig.~\ref{fig:patient_flowchart}.

The study was conducted with institutional authority review board approval from both centers (UW15-596), and written informed consent was obtained from all participants.

\subsection{Model Architecture}

\subsubsection{HRNet Backbone}

We apply High-Resolution Network (HRNet) \cite{wang2020deep} as our backbone for multi-scale feature extraction. It extracted feature maps at four scales (1/4, 1/8, 1/16, and 1/32 resolution). Finally, these four feature maps are fused and passed to the next layer. The feature maps generated in this way retain the advantage of multiple-scale resolutions and have an integrated representation of both local and global features.

\subsubsection{Transformer Enhancement}

To balance the need for high-resolution feature maps and the prohibitive computational cost of traditional self-attention, we adopted the Swin Transformer architecture \cite{liu2021swin}, which implements hierarchical, window-based attention. Unlike global self-attention with $O\left(n^2\right)$ complexity, Swin Transformer reduces computation to $O\left(w\times h\cdot n\right)$, where $w$ and $h$ are window dimensions and $n$ is the sequence length. We employed alternating regular and shifted window attention to enable cross-window information exchange while preserving spatial precision. Multiple attention heads and transformer layers were configured based on empirical performance optimization for vertebral landmark detection \cite{vaswani2017attention}.

\subsubsection{Output Head}

Our prediction targets include both the superior and inferior endplate angles of each vertebra and their corresponding midpoints. We designed an output head that jointly predicts heatmaps for landmark localization and vector fields for angle regression at the same resolution. The heatmaps determine landmark positions and guide the extraction of angle vectors from the corresponding spatial locations \cite{zou2023vltenet}. Inspired by W-Transformer \cite{yao2022w}, we adopted a dual-heatmap strategy that predicts upper and lower endplate midpoints separately, reducing landmark interference while preserving spatial relationships between vertebral elements.

\subsubsection{Biomechanically Informed Loss}

In our model, we explicitly incorporate anatomical knowledge into the loss function to ensure biomechanically plausible predictions. Specifically, we define three essential geometric measures characterizing spinal alignment:
\begin{itemize}[leftmargin=1em]
    \item \textbf{Endplate Tilt Angles $\left( \theta^{\text{upper}},\theta^{\text{lower}} \right) $:} These angles measure the inclination of vertebral endplates relative to the horizontal line, with positive values indicating rightward tilt and negative values indicating leftward tilt.

    \item \textbf{Endplate Midpoints:} These midpoints are employed to calculate directional angles between vertebrae.

    \item \textbf{Directional Angle ($\beta$):} This angle quantifies the orientation between adjacent vertebrae and is defined as:
    \begin{equation}
    \beta_{i,j} = \arctan2(\Delta x_{i,j}, \Delta y_{i,j}),
    \end{equation}
    where $(\Delta x_{i,j}, \Delta y_{i,j})$ represents the vector connecting the geometric centers of vertebra $i$ to vertebra $j$.
\end{itemize}

Our loss function combines three distinct components: the heatmap loss $\ell_{h}$, the vector loss $\ell_{v}$ and the constraint loss $\ell_{c}$,
\begin{equation}
\mathcal{L} = \lambda_1 \ell_{h} + \lambda_2 \ell_{v} + \lambda_3 \ell_{c}.
\end{equation}

The \textbf{heatmap loss} $\ell_{h}$ utilizes a weighted mean squared error to emphasize critical landmarks over background regions:
\begin{equation}
\ell_{h} = N^{-1} \cdot \textstyle\sum_{i}  (\hat{h}_i - h_i)^2 \cdot \tau^{h_i},
\end{equation}
where $\hat{h}_i$ and $h_i$ are predicted and ground truth heatmap values, and $\tau$ is a weighting parameter.

The \textbf{vector loss} $\ell_{v}$ employs mean absolute error, effectively preventing the disproportionate influence of large errors:
\begin{equation}
\ell_{v} = N^{-1} \cdot \textstyle\sum_{i} |\hat{\vec{v}}_i - \vec{v}_i|,
\end{equation}
where $\hat{\vec{v}}_i$ and $\vec{v}_i$ correspond to predicted and ground truth vectors of vertebral endplate angles, respectively.

The \textbf{constraint loss} $\ell_{c}$ constitutes the core of our biomechanical enforcement mechanism, compelling the vertebral tilt angles to conform to natural spinal curvature principles:
\begin{equation}
\ell_{c} = \textstyle\sum\nolimits_{i=1}^{N-2} |\theta_i - \beta_{i}|\cdot\sigma,
\end{equation}
where $\theta_i = (\theta^{\text{upper}}_i + \theta^{\text{lower}}_i) / 2$ represents the average tilt angle of vertebra $i$, and $\beta_{i} = (\beta_{i-1,i} + \beta_{i,i+1})/2$ represents the average directional angle between adjacent vertebrae. This formulation ensures vertebral tilt angles adherence to the fundamental biomechanical constraint, i.e., $|\theta_i - \beta_{i}| \leq \varepsilon_i$. The function $\sigma(\cdot)$ acts as an indicator to ensure the constraint penalizes deviations only when they exceed predefined thresholds, where $\sigma(\cdot) = 1$ if $|\theta_i - \beta_{i}| > \varepsilon_i$ and $\sigma(\cdot) = 0$ otherwise.

This constraint is grounded in spinal biomechanics principles: vertebral tilt angles are naturally constrained within specific ranges due to interconnected ligamentous structures and articular facet orientations. This enables physiologically plausible predictions while retaining flexibility to capture pathological curvatures.

\subsection{Singular Value Decomposition Analysis}
Using our deep learning model, we obtain each vertebra's position and endplate tilt angles. Here, we transform these measurements into diagnostic information.

\subsubsection{Construction of Angle Matrix}
In simplified models, each vertebra is represented by a single tilt angle \cite{zou2023vltenet,lin2020seg4reg, yao2022w, chen2019accurate}. However, when considering both upper and lower endplate angles, we need a more detailed representation. The angle matrix $\Gamma$ defined as:
\begin{equation}
\Gamma = \boldsymbol{\theta}_{upper} \mathbf{1}^{\top} - \mathbf{1} \boldsymbol{\theta}_{lower}^{\top},
\end{equation}

where $\theta_{upper}$ and $\theta_{lower}$ are row vectors of upper and lower endplate angles, and 1 is an 18-dimensional column vector of ones. Each element $\gamma_{i,j}$ represents the angle between the upper endplate of vertebra $i$ and lower endplate of vertebra $j$:
\begin{equation}
\gamma_{i,j} = \theta^{\text{upper}}_i - \theta^{\text{lower}}_j.
\end{equation}

\begin{figure}[!t]
\centering
\includegraphics[width=\columnwidth]{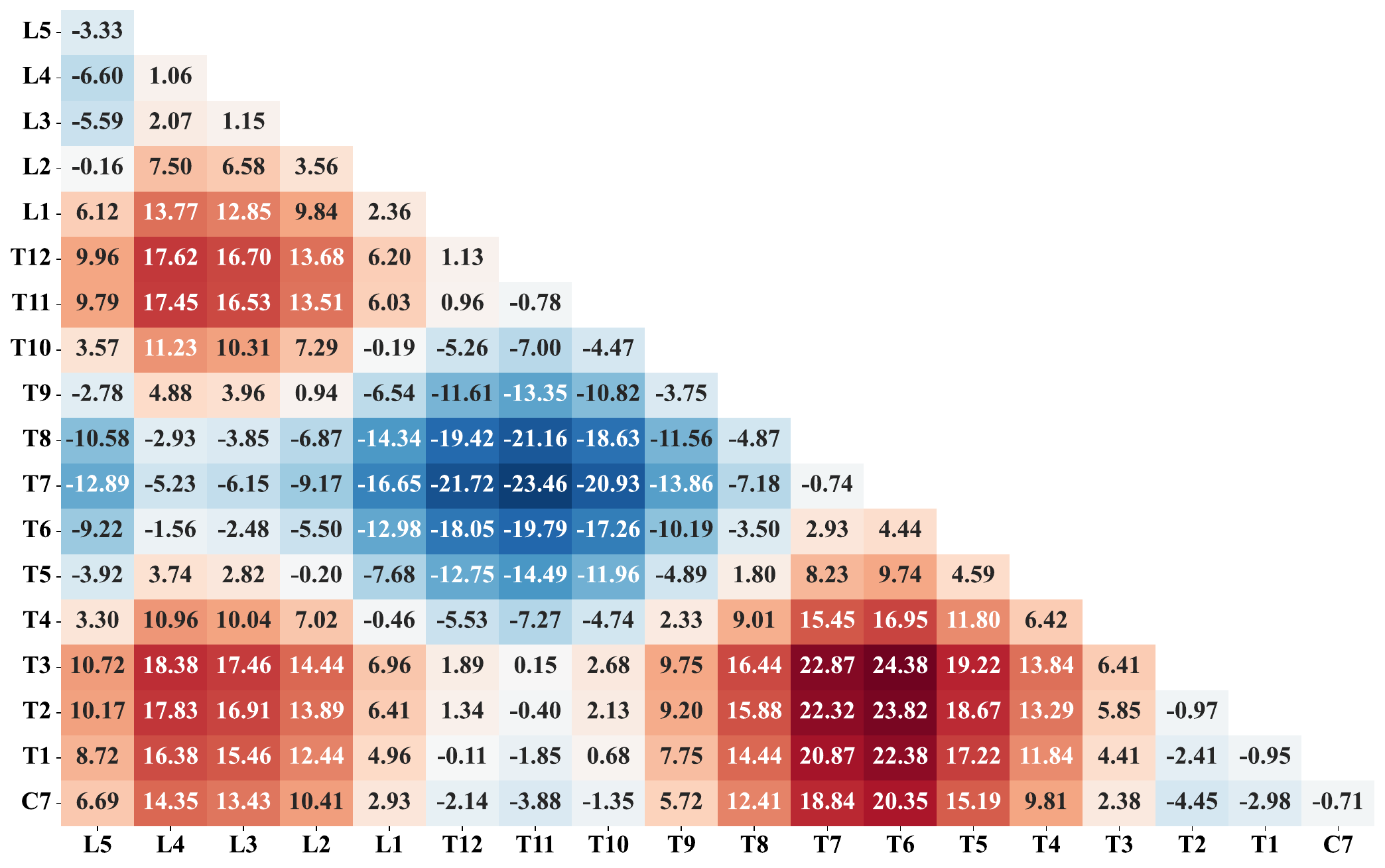}
\caption{Angle matrix $\Gamma$ heatmap showing vertebrae pair angles (in degrees). Diagonal elements represent angles between upper and lower endplates of the same vertebra, while positive/negative values indicate opposite curvature directions. Upper right elements were removed as they represent angles between lower vertebrae's superior endplates and upper vertebrae's inferior endplates, which lack clinical significance.}
\label{fig:angle_matrix_heatmap}
\end{figure}

\subsubsection{SVD for Pattern Extraction}
Dimensionality reduction has proven valuable in medical image analysis, with SVD successfully applied in medical image processing \cite{patil2021medical}.

We apply SVD to decompose the angle matrix:
\begin{equation}
\Gamma = U\Sigma V^T,
\end{equation}
where $U$ and $V$ are orthogonal matrices containing singular vectors, and $\Gamma$ is a diagonal matrix of singular values in descending order.

\subsection{Diagnosis Algorithm}
\begin{figure}[!t]
\centering
\includegraphics[width=\columnwidth]{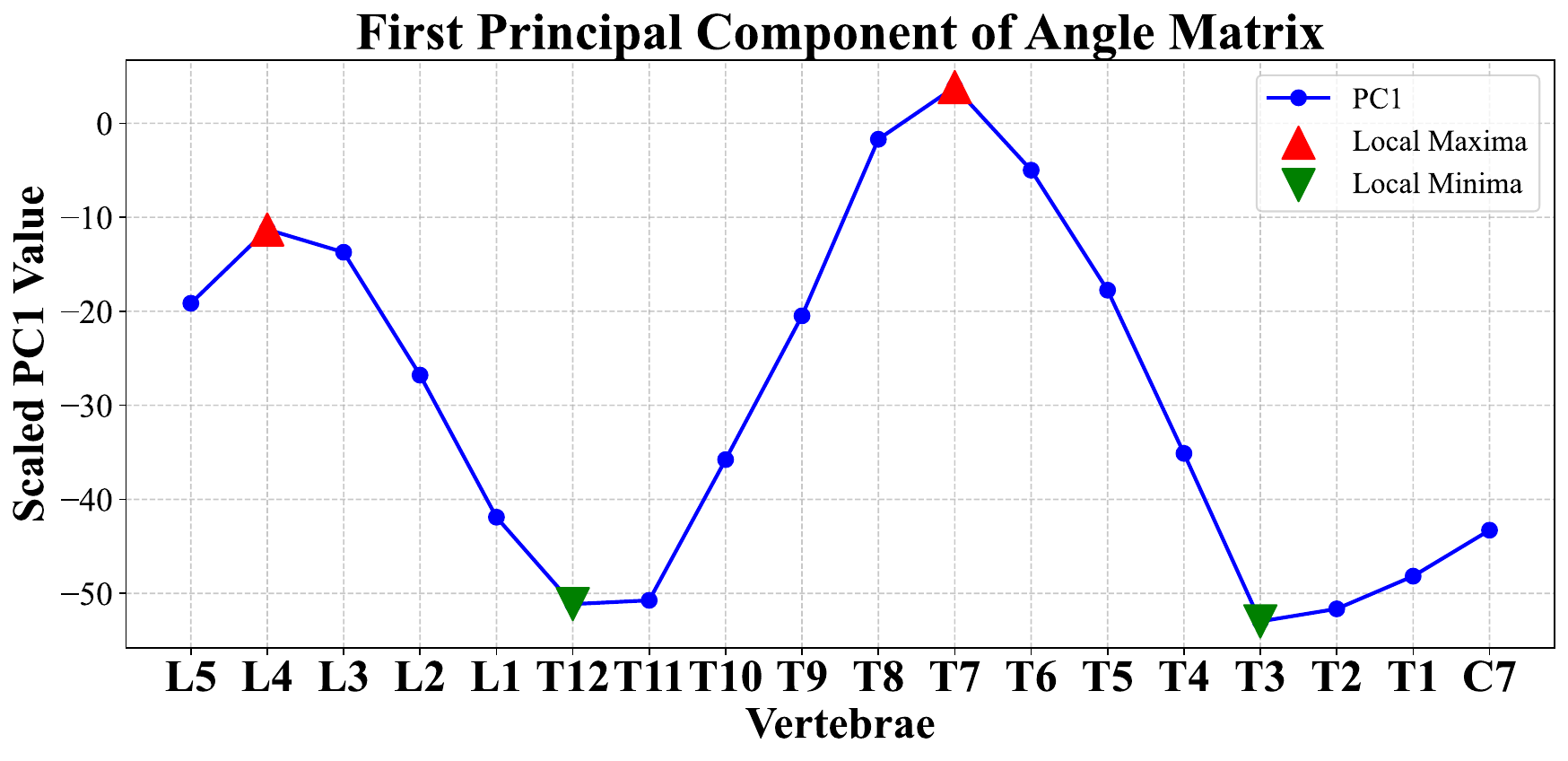}
\caption{Visualization of the first principal component ($PC_1$) curve derived from the angle matrix. The local extrema (peaks and valleys) represent potential end vertebrae of spinal curves, which form the basis for our curve identification algorithm.}
\label{fig:angle_principal_component}
\end{figure}
Building upon the mathematical foundation established through SVD, we propose a principled algorithm to identify clinically meaningful spinal curves and their corresponding end vertebrae. We define a set of potential end vertebrae, $E_{pot}$, by detecting local extrema on the first principal component $PC_1$ curve derived from the angle matrix, $E_{pot} = \left\{ v_i\mid PC_1(v_i) \text{ is a local extrema} \right\}$. For a vertebra $v_i$ to qualify as a local maximum, it must satisfy the following condition:
{\small
\begin{align}
\text{PC1}(v_i) &\geq \text{PC1}(v_{i-2}) \land \text{PC1}(v_i) \geq \text{PC1}(v_{i-1}) \land \nonumber\\
&\text{PC1}(v_i) \geq \text{PC1}(v_{i+1}) \land \text{PC1}(v_i) \geq \text{PC1}(v_{i+2}).
\end{align}
}
A mirrored condition is applied for local minima. The peaks and valleys illustrated in Fig.~\ref{fig:angle_principal_component} correspond to these extrema points. To address anatomical variability in the lumbar region, we exclude the 5th lumber vertebra (L5) due to sacral articulation and apply relaxed criteria for L2–L4. In cases where no clear extrema are detected, L4 may be designated as the terminal vertebra following clinical precedent \cite{lenke2003lenke}. When multiple consecutive vertebrae satisfy the extremum criteria, we select the most representative vertebra to avoid over-segmentation. Clinical significance is determined by an angle threshold:
\begin{equation}
|\gamma_{i,j}| \geq \gamma_{threshold},
\end{equation}
where $\gamma_{threshold} = 10^{\circ}$ \cite{hresko2013idiopathic}. Curve direction is inferred from the sign of the angle matrix value: $\gamma_{i,j}>0$ indicates a rightward curvature (dextroscoliosis), while and $\gamma_{i,j}<0$ indicates a leftward curvature (levoscoliosis). Finally, post-processing steps are employed to refine the diagnostic output. These include removing adjacent curves with overlapping end vertebrae and merging neighboring curves with the same directional polarity. This refinement ensures consistency with clinical practice and enhances diagnostic interpretability.

\subsection{Novel Metric: Vertebral Wedging Index}
\begin{figure}[!t]
    \centering
    \begin{subfigure}[b]{0.38\columnwidth}
        \centering
        \includegraphics[width=\textwidth]{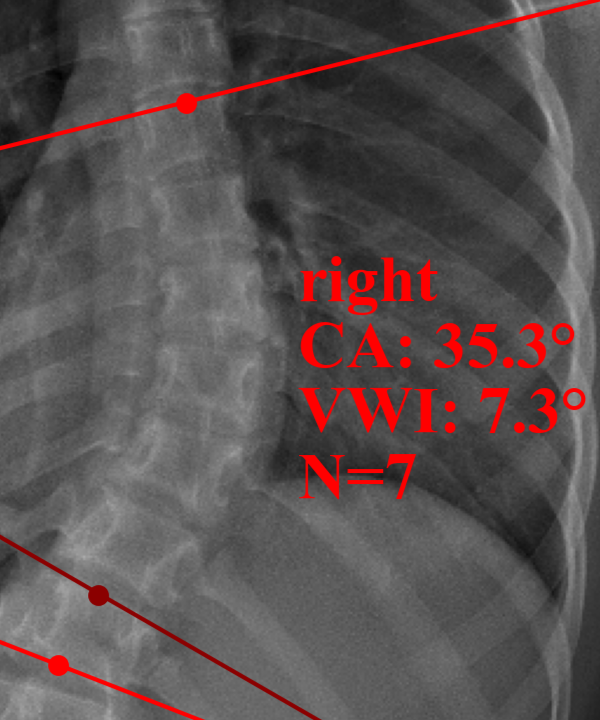}
        \caption{}
        \label{fig:high_vwi}
    \end{subfigure}
    \hspace{0.02\textwidth}
    \begin{subfigure}[b]{0.38\columnwidth}
        \centering
        \includegraphics[width=\textwidth]{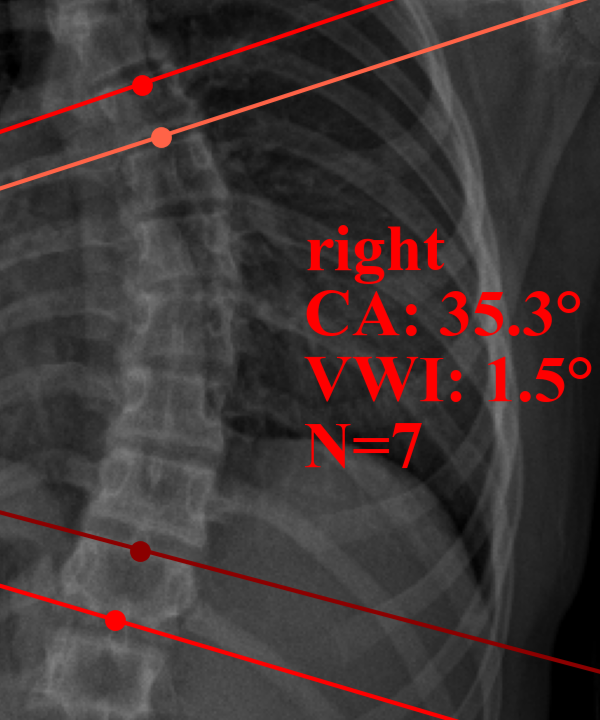}
        \caption{}
        \label{fig:low_vwi}
    \end{subfigure}
    \caption{Cases with similar Cobb angles but different VWI values (1.5° vs 7.3°), demonstrating VWI's ability to differentiate structural characteristics beyond traditional measurements.}
    \label{fig:VWI}
\end{figure}

Clinical research shows that AIS progression involves initial changes to intervertebral discs, followed by structural modifications to vertebral bodies, resulting in wedge-shaped deformities \cite{clemente2012morphological}. We therefore introduce the \textbf{Vertebral Wedging Index (VWI)} to quantify vertebral deformation as illustrated in Fig.~\ref{fig:VWI}, since the degree of vertebral wedging could reflect disease progression characteristics:
\begin{equation}
\text{VWI} = (N)^{-1} \textstyle\sum\nolimits_{i=s}^{s+N-1} |\theta^{\text{upper}}_i - \theta^{\text{lower}}_i|,
\end{equation}
where $s$ is the superior end vertebra index, $N$ is the number of vertebrae in the curve, and $\theta^{\text{upper}}_i$, $\theta^{\text{lower}}_i$ are endplate angles.

\section{Experiments and Results}

\subsection{Evaluation Metrics}
\subsubsection{Clinical Assessment Metrics}
\begin{itemize}[leftmargin=1em]
\item \textbf{Max Cobb Angle Mean Absolute Error(MMAE)}: Average absolute difference calculated between predicted and ground truth maximum Cobb angles across all test cases:
\begin{equation}
\text{MMAE} = N^{-1} \textstyle\sum\nolimits_{i=1}^{N} |\text{MCA}_{\text{pred},i} - \text{MCA}_{\text{GT},i}|,
\end{equation}
where $\text{MCA}_{\text{pred}}$ and $\text{MCA}_{\text{GT}}$ represent the predicted and ground truth maximum Cobb angles respectively.

\item \textbf{Diagnostic Accuracy (DA)}: Proportion of cases correctly classified according to severity thresholds (Normal/Mild: $<20^\circ$, Moderate: $20^\circ-40^\circ$, Severe: $\geq40^\circ$):
\begin{equation}
\text{DA} = N^{-1} \textstyle\sum\nolimits_{i=1}^{N} I(S_{\text{pred},i} = S_{\text{GT},i}) \times 100\%,
\end{equation}
where $I(\cdot)$ is the indicator function and $S_{\text{pred}}$ and $S_{\text{GT}}$ represent the severity classification of the predicted and ground truth maximum Cobb angles.

\item \textbf{Curve Detection Rate (CDR)}: Proportion of curves where both the predicted upper and lower end vertebrae are within $\pm1$ vertebra of their ground truth locations:
\begin{equation}
{\small
\begin{aligned}
\text{CDR} &= N_{\text{GT}}^{-1}\! \textstyle\sum\nolimits_{i=1}^{N_{\text{GT}}}\! I(|\text{UV}_{\text{pred},i} - \text{UV}_{\text{GT},i}|\!\leq\!1 \\
&\quad\land\! |\text{LV}_{\text{pred},i} - \text{LV}_{\text{GT},i}|\!\leq\!1) \times 100\%,
\end{aligned}
}
\end{equation}
where $\text{UV}_{\text{pred}}$ and $\text{UV}_{\text{GT}}$ are the predicted and ground truth upper end vertebrae, and $\text{LV}_{\text{pred}}$ and $\text{LV}_{\text{GT}}$ are the predicted and ground truth lower end vertebrae.

\item \textbf{False Detection Rate (FDR)}: Proportion of predicted curves that do not match any ground truth curve, quantifying the model's specificity in curve identification:
\begin{equation}
{\small
\begin{aligned}
\text{FDR} &= N_{\text{pred}}^{-1}\! \textstyle\sum\nolimits_{j=1}^{N_{\text{pred}}}\! I(\forall i \leq N_{\text{GT}}: |\text{UV}_{\text{pred},j} - \text{UV}_{\text{GT},i}|\\
&\quad\!>\!1 \lor |\text{LV}_{\text{pred},j} - \text{LV}_{\text{GT},i}|\!>\!1) \times 100\%.
\end{aligned}
}
\end{equation}
\end{itemize}

\subsubsection{Model Assessment Metrics}

\begin{itemize}[leftmargin=1em]
\item \textbf{Mean Position Error (MPE)}: The positional accuracy of predicted endplate midpoints:
\begin{equation}
\text{MPE} = 36\cdot(N)^{-1}\textstyle\sum\nolimits_{i=1}^{N}\textstyle\sum\nolimits_{j=1}^{36}{||\mathbf{P}_{i,j}^{\text{pred}} - \mathbf{P}_{i,j}^{\text{GT}}||_2},
\end{equation}
where $\mathbf{P}^{\text{pred}}$ and $\mathbf{P}^{\text{GT}}$ represent the predicted and ground truth points.

\item \textbf{Mean Angle Error (MAE)}: The angular accuracy of predicted vertebral orientations:
\begin{equation}
\text{MAE} = 36\cdot(N)^{-1}\textstyle\sum\nolimits_{i=1}^{N}{\textstyle\sum\nolimits_{j=1}^{36}{||\mathbf{v}_{i,j}^{\text{pred}} - \mathbf{v}_{i,j}^{\text{GT}}||_2}},
\end{equation}
where $\mathbf{v}^{\text{pred}}$ and $\mathbf{v}^{\text{GT}}$ represent the predicted and ground truth directional vectors.
\end{itemize}

To establish the reliability of our findings, we conducted five independent training runs with different random seeds. Performance metrics are reported as the mean value with standard deviations across these five runs.

\subsection{Implementation Details}

All experiments were conducted using PyTorch framework on NVIDIA GPUs. Input spinal radiographs were resized to 1536$\times$512 pixels with standard normalization. The network was trained for 100 epochs using Adam optimizer with an initial learning rate of $1\times10^{-4}$, weight decay of $5\times10^{-4}$, and batch size of 4. We employed learning rate scheduling with warmup and cosine annealing decay. Additional implementation details are available in our public repository at https://github.com/SCBoy1007/AIS-Cobb-SVD-Analysis.

\subsection{Diagnostic Performance}

\subsubsection{Spinal Curve Detection Accuracy}

To validate the diagnostic advantages of our proposed method, we compared it against two state-of-the-art approaches: VLTENet \cite{zou2023vltenet} and Seg4Reg \cite{lin2020seg4reg}. For fair comparison, we adapted both methods to incorporate the same 10° clinical significance threshold used in our approach, eliminating clinically insignificant curves.

\begin{table}[!t]
\caption{Performance comparison of different methods.}
\label{tab:curve_detection_comparison}
\centering
\renewcommand{\arraystretch}{1.3}
\setlength{\tabcolsep}{4pt}
\begin{tabular}{|c|c|c|c|c|c|}
\hline
\textbf{Method}& \textbf{MMAE(°)} & \textbf{DA(\%)} & \textbf{CDR(\%)} & \textbf{FDR(\%)} \\
\hline
VLTENet & 2.89±0.23 & 78.65±3.73& \textbf{98.72±0.73} & 15.89±1.98  \\
\hline
Seg4Reg & 3.24±0.24 & 76.12±3.82  & - & - \\
\hline
Ours & \textbf{2.55±0.20} & \textbf{83.45±3.33} & 97.84±0.90 & \textbf{6.69±1.57}  \\
\hline
\end{tabular}
\end{table}

As shown in Table~\ref{tab:curve_detection_comparison}, our method achieved maximum Cobb angle error of 2.55±0.20° and diagnostic accuracy of 83.45±3.33\%. While VLTENet showed higher curve detection rate (98.72\% vs. 97.84\%), our approach reduced false detections to 6.69\% vs. 15.89\%. Seg4Reg lacks the ability to localize specific curves.

\subsubsection{Disease Severity Classification}

\begin{figure*}[!h]
    \centering
    \begin{subfigure}[b]{0.28\textwidth}
        \centering
        \includegraphics[width=\textwidth]{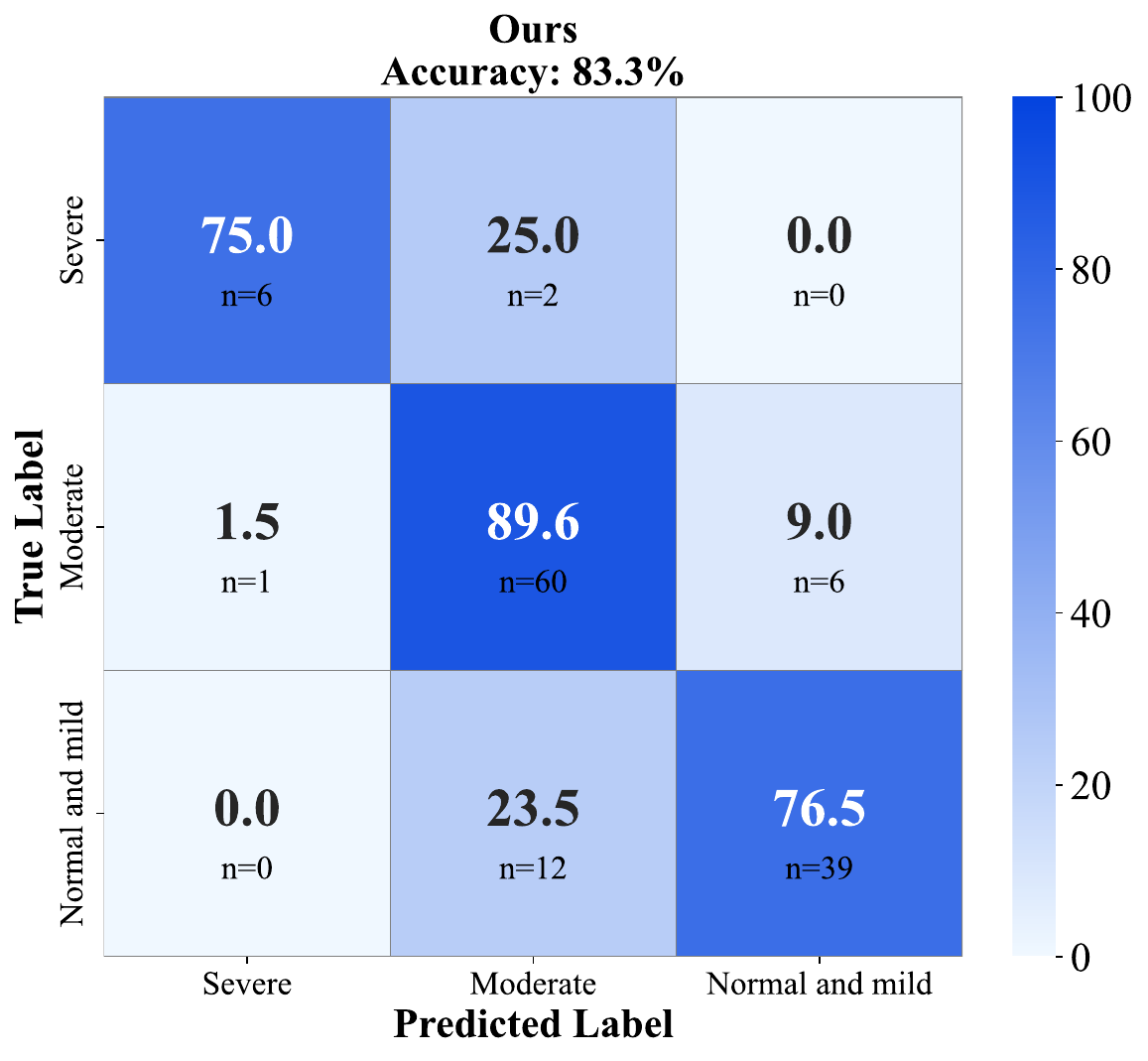}
        \caption{}
        \label{fig:confusion_ours}
    \end{subfigure}
    \hspace{0.01\textwidth}
    \begin{subfigure}[b]{0.28\textwidth}
        \centering
        \includegraphics[width=\textwidth]{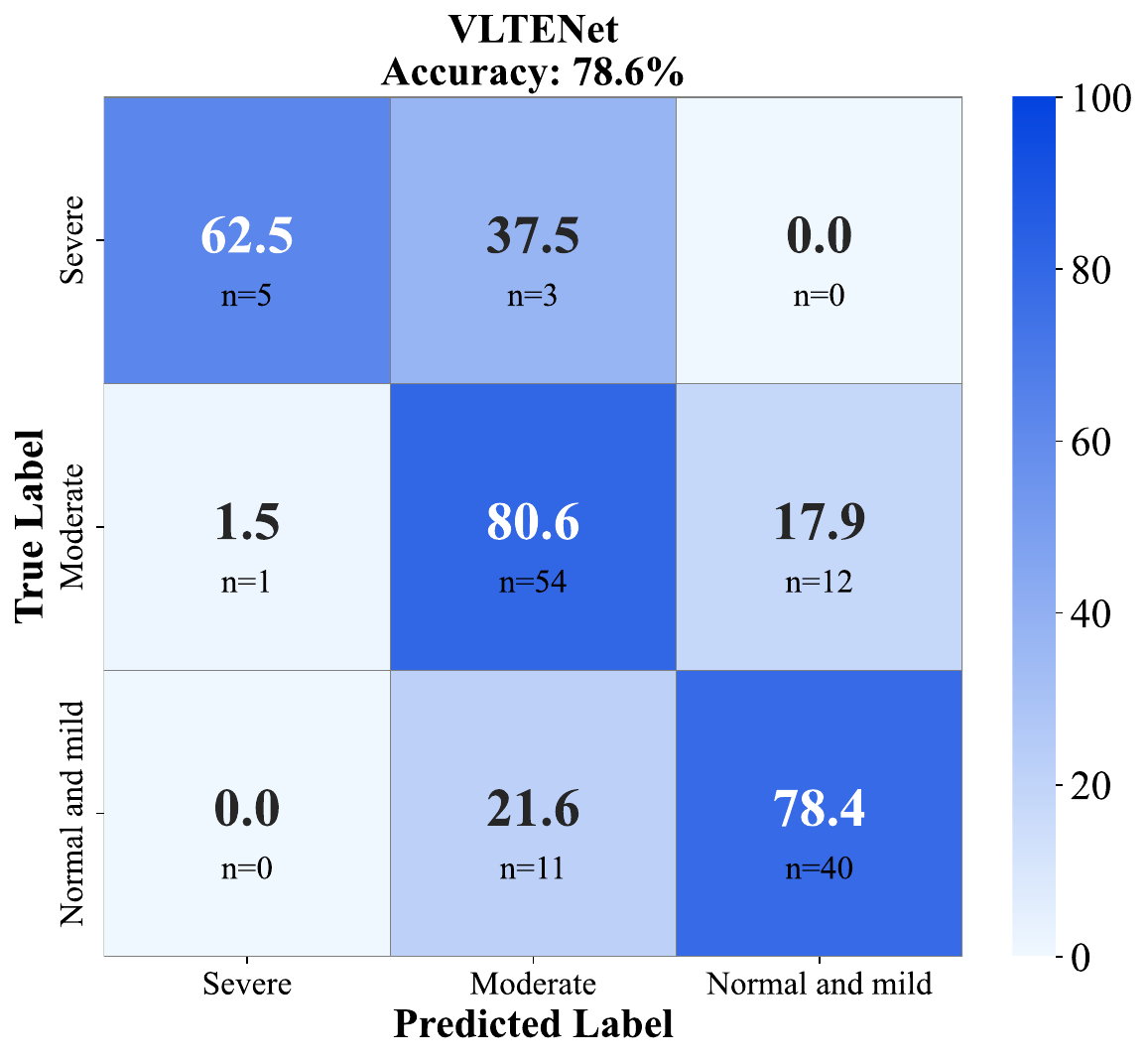}
        \caption{}
        \label{fig:confusion_vltenet}
    \end{subfigure}
    \hspace{0.01\textwidth}
    \begin{subfigure}[b]{0.28\textwidth}
        \centering
        \includegraphics[width=\textwidth]{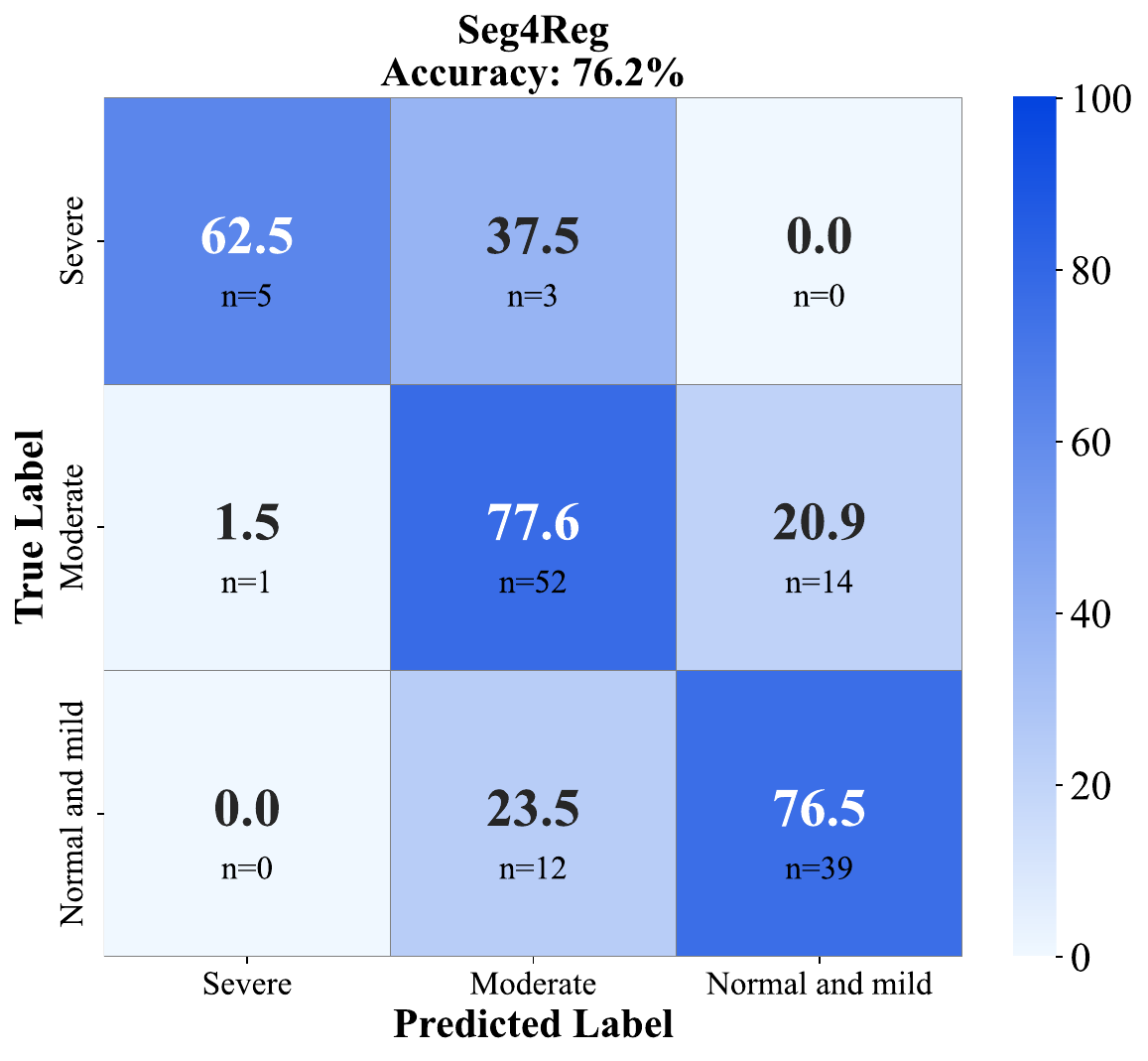}
        \caption{}
        \label{fig:confusion_seg4reg}
    \end{subfigure}
    \caption{Confusion matrices for disease classification. (a), (b) and (c) display the confusion matrix results of our method, VLTNet, and Seg4Reg, respectively. Values are presented as percentages of the total cases in each true category, with actual counts shown in each cell.}
    \label{fig:confusion_matrices_comparison}
\end{figure*}

\begin{table}[!t]
\caption{Classification performance metrics comparison.}
\label{tab:performance_metrics}
\centering
\renewcommand{\arraystretch}{1.2}
\setlength{\tabcolsep}{4pt}
\begin{tabular}{|c|c|c|c|c|}
\hline
\textbf{Category} & \textbf{Method} & \textbf{Precision (\%)} & \textbf{Recall (\%)} & \textbf{F1 (\%)} \\
\hline
\multirow{3}{*}{Severe} & Ours & \textbf{85.71} & \textbf{75.00} & \textbf{80.00} \\
\cline{2-5}
 & VLTENet & 83.33 & 62.50 & 71.43 \\
\cline{2-5}
 & Seg4Reg & 83.33 & 62.50 & 71.43 \\
\hline
\multirow{3}{*}{Moderate} & Ours & \textbf{81.08} & \textbf{89.55} & \textbf{85.11} \\
\cline{2-5}
 & VLTENet & 79.41 & 80.60 & 80.00 \\
\cline{2-5}
 & Seg4Reg & 77.61 & 77.61 & 77.61 \\
\hline
\multirow{3}{*}{Normal/Mild} & Ours & \textbf{86.67} & 76.47 & \textbf{81.25} \\
\cline{2-5}
 & VLTENet & 76.92 & \textbf{78.43} & 77.67 \\
\cline{2-5}
 & Seg4Reg & 73.58 & 76.47 & 75.00 \\
\hline
\end{tabular}
\end{table}

As shown in Fig.~\ref{fig:confusion_matrices_comparison} and Table~\ref{tab:performance_metrics}, our method achieved overall classification accuracy of 83.33\% (vs. 78.57\% for VLTENet and 76.19\% for Seg4Reg). In the Severe category, recall improved to 75.00\% vs. 62.50\%. In the Moderate category, F1 score reached 85.11\% with recall of 89.55\%. For Normal/Mild cases, precision was 86.67\% and F1 score was 81.25\%.

\begin{figure*}[!h]
    \centering
    \begin{subfigure}[b]{0.25\textwidth}
        \centering
        \includegraphics[width=\textwidth]{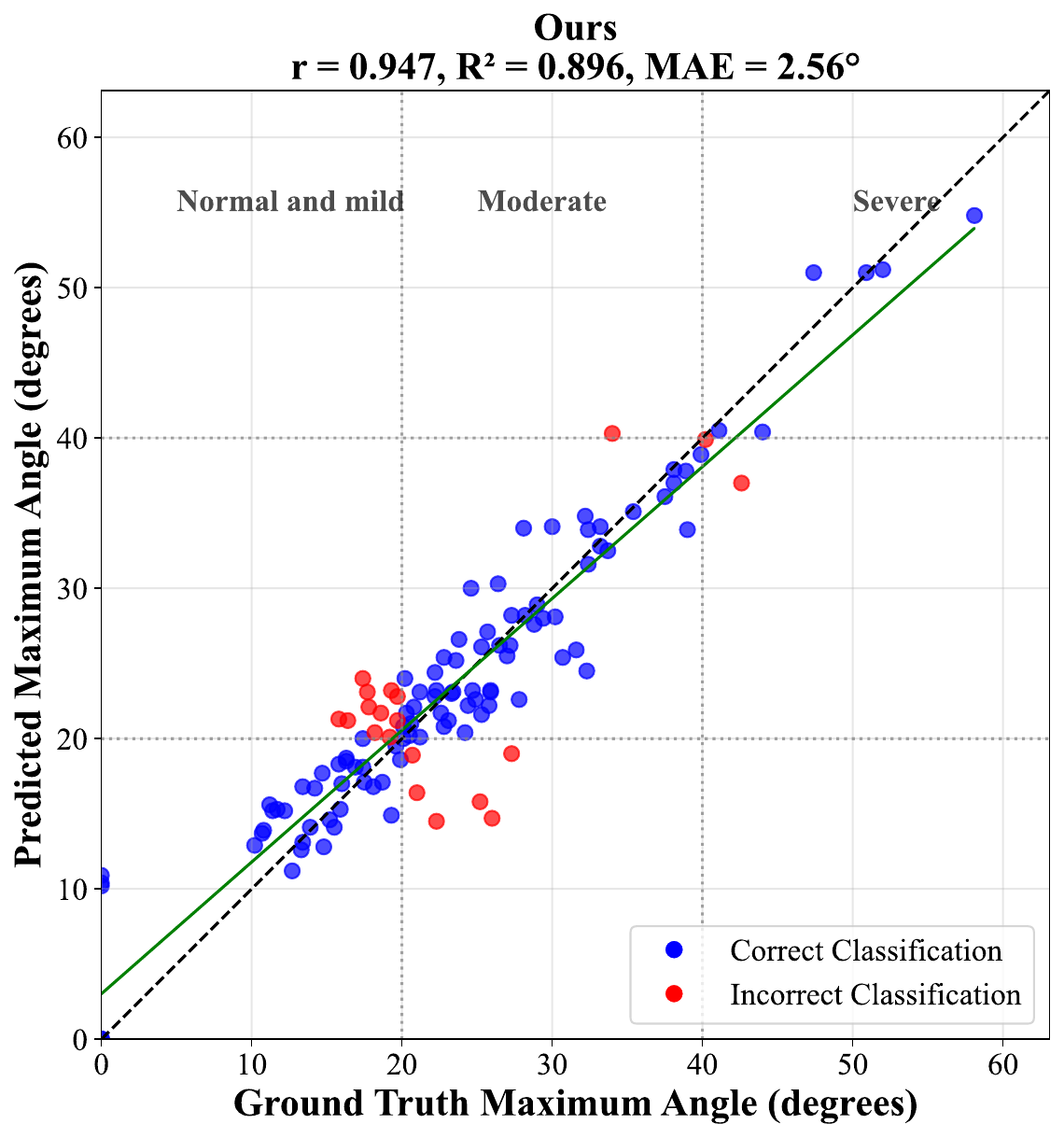}
        \caption{}
        \label{fig:correlation_ours}
    \end{subfigure}
    \hspace{0.05\textwidth}
    \begin{subfigure}[b]{0.25\textwidth}
        \centering
        \includegraphics[width=\textwidth]{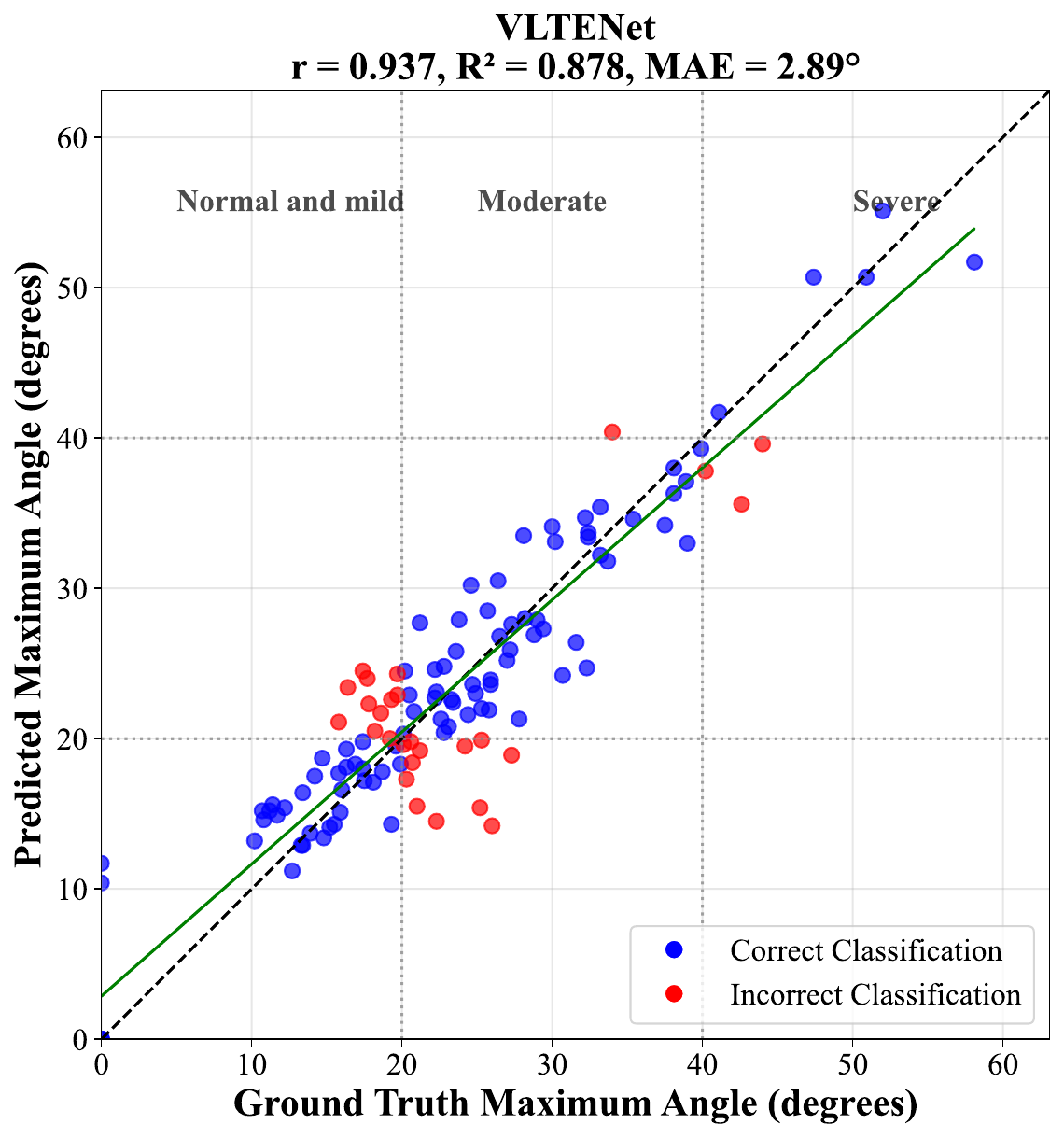}
        \caption{}
        \label{fig:correlation_vltenet}
    \end{subfigure}
    \hspace{0.04\textwidth}
    \begin{subfigure}[b]{0.25\textwidth}
        \centering
        \includegraphics[width=\textwidth]{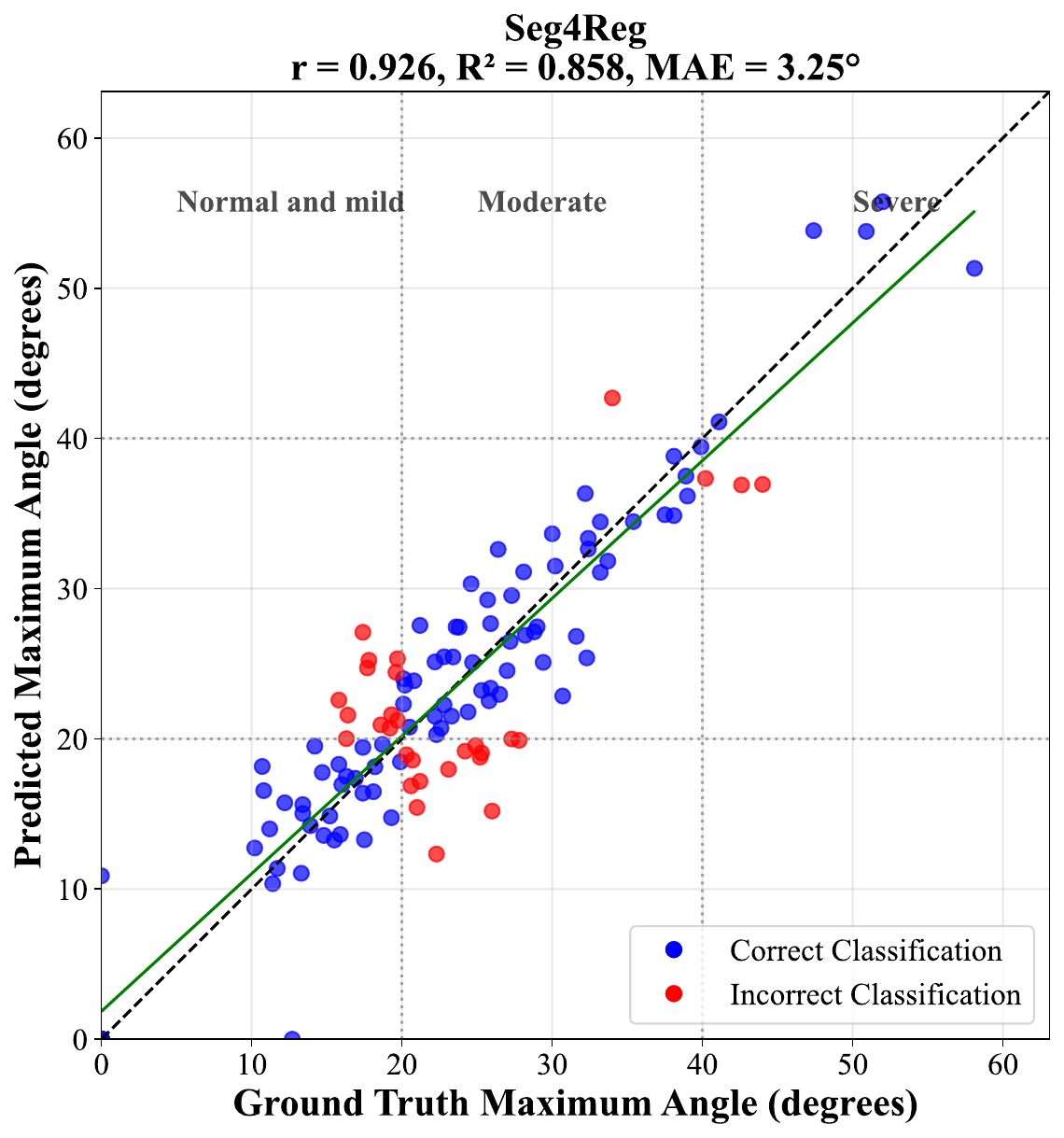}
        \caption{}
        \label{fig:correlation_seg4reg}
    \end{subfigure}
    \hspace{0.03\textwidth}
    \caption{Correlation between ground truth and predicted maximum Cobb angles.
(a) Our method, (b) VLTENet, and (c) Seg4Reg. Blue: correct classifications; red: misclassifications. Grid lines at 20° and 40° indicate clinical thresholds.}
    \label{fig:correlation_analysis}
\end{figure*}

Fig.~\ref{fig:correlation_analysis} presents scatter plots comparing ground truth and predicted maximum Cobb angles. Misclassifications predominantly occur near clinical decision thresholds (20° and 40°), which is clinically acceptable given the 5°-10° measurement variability in manual assessment. Our method achieved correlation (r = 0.947, R² = 0.896, MAE = 2.56°) compared to VLTENet (r = 0.937, R² = 0.878, MAE = 2.89°) and Seg4Reg (r = 0.926, R² = 0.858, MAE = 3.25°). 80.00\% of misclassifications occurred within 5° of classification boundaries, with 20 total classification errors vs. 27 and 30 cases for comparison methods.

\subsection{Ablation Study}

To systematically evaluate our design choices and identify the key contributors to model performance, we conducted three sets of ablation experiments.

\subsubsection{Backbone Selection}

\begin{table}[!t]
\caption{Performance Comparison of Different Backbones.}
\label{tab:backbone-comparison}
\centering
\renewcommand{\arraystretch}{1.3}
\setlength{\tabcolsep}{4pt}
\begin{tabular}{|c|c|c|c|}
\hline
\textbf{Backbone} & \textbf{MPE(px)} & \textbf{MAE(°)} & \textbf{Params}  \\
\hline
HRNet-18 & 4.74±0.80 & 2.10±0.08 & 9.56M  \\
\hline
HRNet-32 & 4.76±1.05 & 2.13±0.10 & 29.31M  \\
\hline
ResNet-50 & 5.74±2.10 & 2.16±0.10 & 25.75M  \\
\hline
ResNet-101 & 6.01±1.99 & 2.21±0.11 & 44.74M  \\
\hline
UNet & 5.30±1.33 & 2.17±0.11 & 7.76M  \\
\hline
UNet++ & 4.84±1.28 & 2.15±0.10 & 8.73M  \\
\hline
DenseNet-121 & 4.83±0.83 & 2.16±0.09 & 10.51M  \\
\hline
DenseNet-169 & 4.94±1.00 & 2.16±0.10 & 21.56M  \\
\hline
EfficientNet-B0 & 5.35±0.95 & 2.25±0.09 & 9.40M \\
\hline
EfficientNet-B3 & 4.83±0.69 & 2.18±0.08 & 18.45M \\
\hline
\end{tabular}
\end{table}

To identify the optimal feature extractor, we evaluated ten representative CNN architectures across five backbone families (Table~\ref{tab:backbone-comparison}). HRNet-18 emerged as the most effective backbone, achieving the lowest angular error (2.10°) while maintaining a moderate parameter count (9.56M). Notably, larger models within the same family did not yield performance improvements despite their increased complexity.

\subsubsection{Output Head Design Evaluation}

To verify the necessity of the two components of our model's output head——a dual heatmap and a vector map, we conducted experiments with alternative approaches: 1) Vector-Only approach, which directly regresses angles from vector map; 2) Heatmap-Only approach, which expands our dual heatmaps to four separate heatmaps and calculates angles indirectly from predicted landmarks; and 3) our proposed hybrid approach (Table~\ref{tab:output_head}).

\begin{table}[!t]
\caption{Performance Comparison of Different Designs.}
\label{tab:output_head}
\centering
\renewcommand{\arraystretch}{1.3}
\setlength{\tabcolsep}{4pt}
\begin{tabular}{|c|c|c|}
\hline
\textbf{Method} & \textbf{MPE(px)} & \textbf{MAE(°)}\\
\hline
Vector-Only & - & 12.72±2.68 \\
\hline
Heatmap-Only& 3.80±0.77 & 3.78±0.23 \\
\hline
Heatmap+Vector & 4.74±0.80 & 2.10±0.08 \\
\hline
\end{tabular}
\end{table}

The Vector-Only approach performed poorly, confirming that spatial context is essential for accurate angle prediction. The Heatmap-Only method achieved better positional accuracy (3.80px vs. 4.74px), but it produced significantly higher angular errors (3.78° vs. 2.10°) than our hybrid approach.

\subsubsection{Component Contribution}
\begin{table}[!t]
\caption{Impact of Different Model Components.}
\label{tab:component-ablation}
\centering
\renewcommand{\arraystretch}{1.3}
\setlength{\tabcolsep}{3pt}
\begin{tabular}{|c|c|c|c|c|}
\hline
\textbf{B} & \textbf{T} & \textbf{$\lambda_1:\lambda_2:\lambda_3$} & \textbf{MPE(px)} & \textbf{MAE(°)} \\
\hline
\ding{51} & \ding{51} & \textbf{1:0.05:0.05} & \textbf{4.74±0.80} & \textbf{2.10±0.08} \\
\hline
\ding{51} & \ding{51} & 1:0.1:0.1 & 4.76±0.82 & 2.12±0.09 \\
\hline
\ding{51} & \ding{51} & 1:0.5:0.5 & 6.23±0.95 & 2.41±0.15 \\
\hline
\ding{55} & \ding{51} & 1:0.05:- & 4.98±1.03 & 2.19±0.11 \\
\hline
\ding{51} & \ding{55} & 1:0.05:0.05 & 5.93±0.87 & 2.17±0.09 \\
\hline
\ding{55} & \ding{55} & 1:0.05:- & 5.86±0.92 & 2.20±0.12  \\
\hline
\end{tabular}
\begin{tablenotes}\footnotesize
\item B: Biomechanically Informed Loss, T: Transformer Enhancement
\item $\lambda_1$, $\lambda_2$, $\lambda_3$ correspond to heatmap, vector, and constraint loss weights
\end{tablenotes}
\end{table}

As shown in Table~\ref{tab:component-ablation}, the Transformer component provided substantial improvement to positional accuracy, reducing MPE by approximately 20\% (from 5.93px to 4.74px). Meanwhile, the anatomical constraints had a more subtle but consistent impact on angular accuracy (improving MAE from 2.20° to 2.10°). The full model combining both components achieved the best overall performance. 
We systematically investigated different weight configurations and used $\lambda_1=1.0$, $\lambda_2=0.05$, $\lambda_3=0.05$ for optimal performance.

\subsection{Representative Cases}
\begin{figure*}[!t]
    \centering
    \begin{subfigure}[t]{0.136\textwidth}
        \centering
        \includegraphics[width=\textwidth]{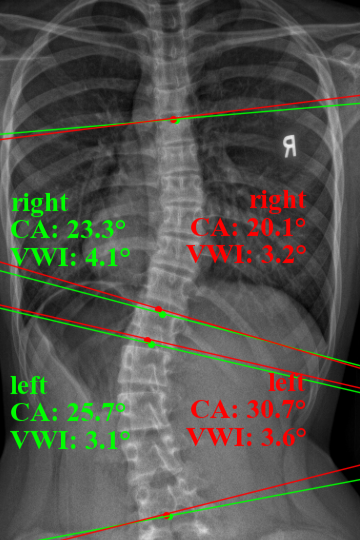}
        \caption{}
        \label{fig:case_A}
    \end{subfigure}
    \hspace{0.02\textwidth}
    \begin{subfigure}[t]{0.136\textwidth}
        \centering
        \includegraphics[width=\textwidth]{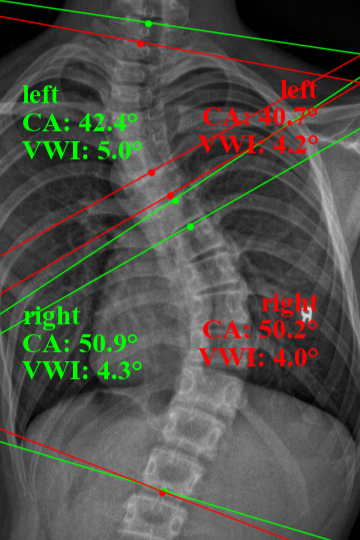}
        \caption{}
        \label{fig:case_B}
    \end{subfigure}
    \hspace{0.02\textwidth}
    \begin{subfigure}[t]{0.136\textwidth}
        \centering
        \includegraphics[width=\textwidth]{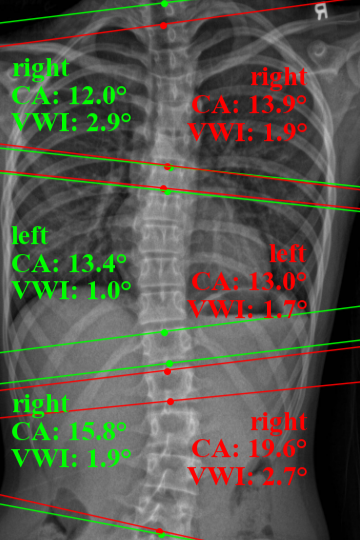}
        \caption{}
        \label{fig:case_C}
    \end{subfigure}
    \hspace{0.02\textwidth}
    \begin{subfigure}[t]{0.136\textwidth}
        \centering
        \includegraphics[width=\textwidth]{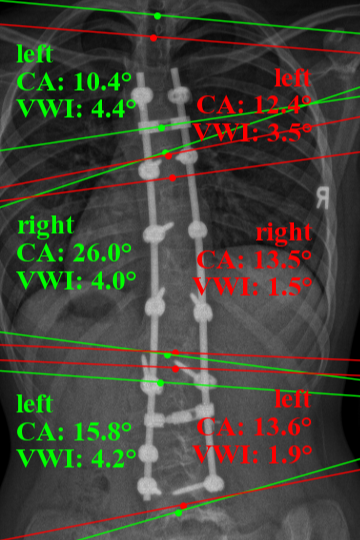}
        \caption{}
        \label{fig:case_D}
    \end{subfigure}
    \hspace{0.02\textwidth}
    \begin{subfigure}[t]{0.136\textwidth}
        \centering
        \includegraphics[width=\textwidth]{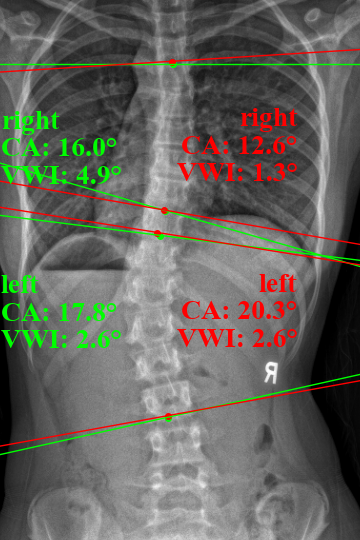}
        \caption{}
        \label{fig:case_E}
    \end{subfigure}
    \hspace{0.02\textwidth}
    \begin{subfigure}[t]{0.115\textwidth}
        \centering
        \includegraphics[width=\textwidth]{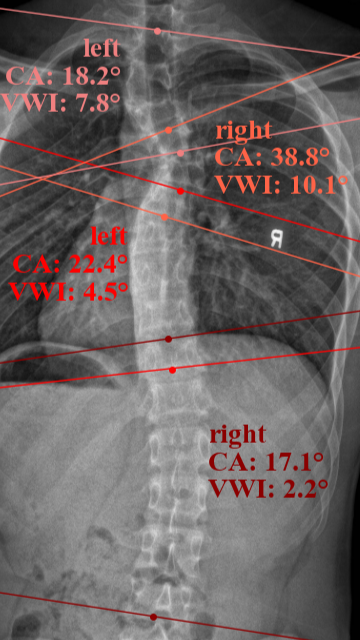}
        \caption{}
        \label{fig:case_F}
    \end{subfigure}
    \caption{Representative cases with different patterns: (a) major lumbar curve, (b) severe major thoracic curve, (c) triple-curve pattern with mild angles, (d) moderate lumbar curve with post-surgical hardware interference, (e) mild double curve with severity overestimation, and (f) four-curve pattern excluded from our dataset. Green lines and text represent ground truth annotations, while red lines and text show our model's predictions.}
    \label{fig:diagnostic_examples}
\end{figure*}

\begin{table}[!t]
\caption{Ground truth and model predictions}
\label{tab:gt_pred_comparison}
\centering
\footnotesize
\setlength{\tabcolsep}{3pt}
\renewcommand{\arraystretch}{1.2}
\begin{tabular}{|c|c|c|c|c|c|c|c|c|c|c|}
\hline
\textbf{} & \multicolumn{2}{c|}{\textbf{Case a}} & \multicolumn{2}{c|}{\textbf{Case b}} & \multicolumn{2}{c|}{\textbf{Case c}} & \multicolumn{2}{c|}{\textbf{Case d}} & \multicolumn{2}{c|}{\textbf{Case e}} \\
\hline
 & GT & Pred & GT & Pred & GT & Pred & GT & Pred & GT & Pred \\
\hline
CS1 & T6 & T6 & C7 & T1 & T1 & T2 & T6 & T7 & T5 & T5 \\
CE1 & T12 & T12 & T7 & T6 & T7 & T7 & L1 & L1 & T10 & T10 \\
CD1 & right & right & left & left & right & right & right & right & right & right \\
CA1 & 23.3° & 20.1° & 42.4° & 40.7° & 12.0° & 13.9° & 26.0° & 13.5° & 16.0° & 12.6° \\
\hline
CS2 & T12 & T12 & T7 & T6 & T7 & T7 & T2 & T3 & T10 & T10 \\
CE2 & L4 & L4 & L2 & L2 & T12 & L1 & T6 & T7 & L3 & L3 \\
CD2 & left & left & right & right & left & left & left & left & left & left \\
CA2 & 25.7° & 30.7° & 50.9° & 50.2° & 13.4° & 13.0° & 10.4° & 12.4° & 17.8° & 20.3° \\
\hline
CS3 & - & - & - & - & T12 & L1 & L1 & L1 & - & - \\
CE3 & - & - & - & - & L4 & L4 & L4 & L4 & - & - \\
CD3 & - & - & - & - & right & right & left & left & - & - \\
CA3 & - & - & - & - & 15.8° & 19.6° & 15.8° & 13.6° & - & - \\
\hline
Class. & M & M & S & S & N & N & M & N & N & M \\
\hline
\end{tabular}
\begin{tablenotes}
\footnotesize
\item \textit{Note}: CS = Curve Start; CE = Curve End; CD = Curve Direction; CA = Curve Angle; N = Normal/Mild; M = Moderate; S = Severe
\end{tablenotes}
\end{table}

Six representative cases are shown in Fig.~\ref{fig:diagnostic_examples} and Table~\ref{tab:gt_pred_comparison}. Our model accurately identified curve patterns in the first three cases, with end vertebrae localization errors consistently within ±1 vertebral level. To systematically assess model performance on excluded cases, we conducted a blinded expert evaluation by two senior spine surgeons on all 104 initially excluded patients. The agreement rates between our model and the first expert evaluator was 78.8\% (83/104), while the agreement with the second expert evaluator was 82.7\% (86/104), demonstrating diagnostic accuracy comparable to our main AIS cohort results.

\subsection{VWI Prognostic Value Analysis}
To evaluate VWI's prognostic potential, we analyzed longitudinal data from 138 patients with multiple follow-up examinations (mean follow-up: 5.2 ± 2.7 months), investigating correlations between baseline measurements and subsequent Cobb angle changes.
\begin{table}[!t]
\caption{Prognostic correlation analysis.}
\label{tab:vwi_prognostic}
\centering
\renewcommand{\arraystretch}{1.3}
\setlength{\tabcolsep}{8pt}

\begin{tabular}{|c|c|c|c|}
\hline
\textbf{Metric} & \textbf{Correlation (r)} & \textbf{p-value} & \textbf{Sig.} \\
\hline
Initial VWI & -0.1925 & \textless0.0001 & Yes \\
\hline
Initial Risser Score & -0.1190 & 0.0067 & Yes \\
\hline
Initial Cobb Angle & 0.0847 & 0.3201 & No \\
\hline
\end{tabular}

\end{table}
As shown in Table~\ref{tab:vwi_prognostic}, initial VWI showed significant negative correlation with Cobb angle progression (r = -0.1925, p \textless 0.0001), while traditional Cobb angles showed no significant correlation with progression (r = 0.0847, p = 0.3201). Risser score demonstrated weaker correlation (r = -0.1190, p = 0.0067) compared to VWI.

\section{Discussion and Conclusion}

When compared to Seg4Reg's black-box design, our approach provides clinical interpretability and achieves Cobb angle accuracy of 2.55° vs 3.24°. Two fundamental innovations distinguish our method from VLTENet: First, our dual-endplate modeling captures the anatomical reality of vertebral wedging in progressive AIS, while VLTENet's single-angle simplification fails to represent these clinically significant deformities. Our approach achieved measurement precision of 2.55° vs 2.89° and diagnostic accuracy of 83.45\% vs 78.65\%. Second, our SVD-based detection algorithm identifies intrinsic curvature patterns without predetermined constraints, compared to VLTENet's rigid three-curve approach. Though both methods achieve high detection rates (97.84\% vs 98.72\%), our approach reduced false positives to 6.69\% vs 15.89\%. Beyond typical AIS patterns, our model demonstrated robust generalizability to excluded cases, achieving 78.8\% and 82.7\% agreement with independent expert evaluators on 104 out-of-distribution patients, supporting the clinical flexibility of our approach.

Our ablation studies yield several valuable insights into efficient model design for AIS assessment. Counterintuitively, increasing model capacity (e.g., from HRNet-18 to HRNet-32) resulted in slightly degraded performance despite the parameter count tripling. This suggests that excessive parameters may increase the risk of overfitting rather than enhancing generalization capability. Regarding output head design, our dual heatmap + vector map approach reduced MAE by approximately 45\% (1.68°) compared to the heatmap-only method, at the cost of a 24\% increase in MPE (1.06px). Our hybrid approach directly predicts angles through the vector map, bypassing error accumulation from indirect angle calculation. The Swin-Transformer components capture long-range dependencies for global spinal structure understanding, while anatomical constraints integrate domain-specific knowledge to improve clinical relevance. Despite these advances, our model exhibits limitations in post-surgical cases where metallic spinal instrumentation interferes with vertebral landmark detection, as demonstrated in Fig.~\ref{fig:case_D}. This represents a known challenge in medical image analysis that warrants future investigation.

Our study introduces VWI as a novel metric that captures structural deformities often overlooked by traditional measurements. Our longitudinal analysis of 138 patients demonstrates VWI's prognostic value, showing significant negative correlation with Cobb angle progression (r = -0.1925, p \textless 0.0001). This enables clinicians to differentiate stable from progressive deformities for appropriate monitoring strategies.

A complex congenital scoliosis case (Fig.~\ref{fig:case_F}) demonstrates VWI's clinical utility. Despite a moderate Cobb angle (38.8°) that typically would not warrant surgical intervention, this patient required operative treatment based on expert assessment. The curve segment had a VWI of 10.1°, substantially higher than typical values, revealing information missed by Cobb angle alone. These findings suggest VWI has potential as a complementary metric to guide treatment decisions between conservative management and surgical intervention, particularly in cases where Cobb angle measurements alone might suggest inappropriate treatment strategies.

\bibliographystyle{IEEEtran}
\bibliography{references}

\end{document}